\documentclass[10pt,journal,compsoc]{IEEEtran}
\ifCLASSOPTIONcompsoc
\else
\fi

\usepackage{lipsum}
\usepackage{amsmath}
\usepackage{bbm}
\usepackage{booktabs} 
\usepackage{graphicx}
\usepackage{color}
\usepackage{float}
\usepackage{array}
\usepackage{multirow}
\usepackage{amssymb}
\usepackage[OT1]{fontenc}

\begin{document}

\title{PointINS: Point-based Instance Segmentation}

\author{Lu Qi$^\ddagger$\thanks{$^\ddagger$Part of the work is done during authors' internship in MEGVII Technology. They share the equal contributions. },
		Yi Wang$^\ddagger$,
		Yukang Chen$^\ddagger$,
	    Yingcong Chen,
	    Xiangyu Zhang$^*$ \thanks{ * indicates the corresponding author.},
	    Jian Sun,
	    Jiaya Jia

\IEEEcompsocitemizethanks{
\IEEEcompsocthanksitem L. Qi, Y. Wang, Y.K Chen and J. Jia are with the Department of Computer Science and Engineering, The Chinese University of Hong Kong.
\IEEEcompsocthanksitem Y.C Chen is currently a postdoctoral associate at the Computer Science and Artiﬁcial Intelligence Lab (CSAIL), Massachusetts Institute of Technology.
\IEEEcompsocthanksitem X. Zhang and J. Sun are with MEGVII Technology.}
}

\definecolor{mypink3}{cmyk}{0, 0.7808, 0.4429, 0.1412}


\IEEEtitleabstractindextext{%
\begin{abstract}
In this paper, we explore the mask representation in instance segmentation with Point-of-Interest (PoI) features. Differentiating multiple potential instances within a single PoI feature is challenging, because learning a high-dimensional mask feature for each instance using vanilla convolution demands a heavy computing burden. To address this challenge, we propose an instance-aware convolution. It decomposes this mask representation learning task into two tractable modules as instance-aware weights and instance-agnostic features. The former is to parametrize convolution for producing mask features corresponding to different instances, improving mask learning efficiency by avoiding employing several independent convolutions. Meanwhile, the latter serves as mask templates in a single point. Together, instance-aware mask features are computed by convolving the template with dynamic weights, used for the mask prediction. 
Along with instance-aware convolution, we propose PointINS, a simple and practical instance segmentation approach, building upon dense one-stage detectors. Through extensive experiments, we evaluated the effectiveness of our framework built upon RetinaNet and FCOS. PointINS in ResNet101 backbone achieves a 38.3 mask mean average precision (mAP) on COCO dataset, outperforming existing point-based methods by a large margin. It gives a comparable performance to the region-based Mask R-CNN \cite{DBLP:conf/iccv/HeGDG17} with faster inference.
\end{abstract}

\begin{IEEEkeywords}
Instance Segmentation, Single-Point Feature.
\end{IEEEkeywords}}

\maketitle

\IEEEdisplaynontitleabstractindextext
\IEEEpeerreviewmaketitle

\IEEEraisesectionheading{\section{Introduction}}
Instance segmentation aims to detect all objects from images at the pixel-level, and involves both object detection~\cite{girshick2014rich,DBLP:conf/iccv/Girshick15,DBLP:conf/nips/RenHGS15,DBLP:conf/cvpr/LinDGHHB17,DBLP:conf/nips/DaiLHS16} and semantic segmentation~\cite{long2015fully,zhao2017pyramid,chen2017deeplab}. Instance segmentation is a critical task in computer vision research community, and plays an indispensable role in a variety of real-world applications, such as autonomous vehicles~\cite{qi2019amodal}, robotics~\cite{morrison2018cartman}, video surveillance~\cite{shu2014human}, etc. Considering its both academic and industrial values, increasing the effectiveness and efficiency of instance segmentation is an important challenge.

In formulation, instance segmentation requires the semantics, identities, and pixel-level locations of objects. Existing studies~\cite{DBLP:conf/iccv/HeGDG17,DBLP:journals/corr/abs-1803-01534,huang2019mask,wang2019solo,chen2019tensormask} demonstrate that learning a proper \emph{mask representation} for characterizing objects' identities and locations remains a vital and open problem. Most state-of-the-art (SOTA) instance segmentation methods~\cite{DBLP:conf/iccv/HeGDG17,DBLP:journals/corr/abs-1803-01534,chen2019hybrid,huang2019mask,lee2020centermask,peng2020snake} construct mask representations using Region-of-Interests (RoIs) features. The most representative method is Mask R-CNN~\cite{DBLP:conf/iccv/HeGDG17}, a region-based convolutional neural network that considers instance masks as refined forms from the corresponding bounding boxes. 
Specifically, it extracts RoIs features from regions containing potential instances~\cite{DBLP:conf/iccv/Girshick15,DBLP:conf/nips/RenHGS15,DBLP:conf/cvpr/LinDGHHB17} and then maps these features to detect objects' bounding boxes and to segment instances' masks. Although RoI features are sufficiently expressive to represent regions for final mask prediction, to obtain RoI features is a complicated process.

\begin{figure}[t]
	\begin{center}
		\includegraphics[width=0.9\linewidth]{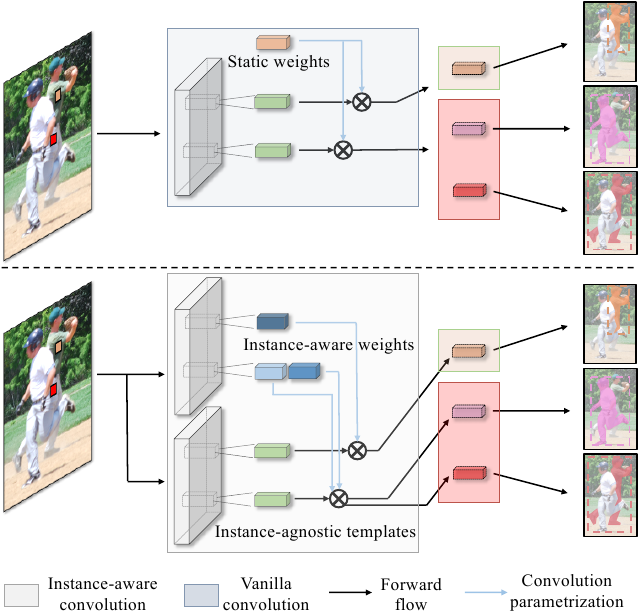}
	\end{center}
	\caption{The core module instance-aware convolution of the proposed PointINS framework in instance segmentation, along with the comparison with the vanilla convolution. By introducing the multiple instance-aware weights from the same point-of-interest (PoI) feature, we obtain different instance masks by instance-aware convolution.}
	\label{fig:sketch}
\end{figure}

To achieve a simple and effective mask representation, several recent approaches, such as TensorMask~\cite{chen2019tensormask}, PolarMask~\cite{xie2020polarmask} and MEInst~\cite{zhang2020MEInst}, directly use single-point features, also known as points-of-interests (PoIs), to learn objects' masks. Based on the assumption that \emph{each single-point feature corresponds to only one or two potential instances}, these approaches give a mask representation from each PoI by the \emph{vanilla convolution} (top row of Fig.~\ref{fig:sketch}), building upon dense one-stage detectors, such as \textit{e.g. RetinaNet~\cite{lin2017focal} or FCOS~\cite{tian2019fcos}}. 

Despite their competitive efficiency in instance segmentation, their performance and scalability are restricted by the adopted assumption and design. In contrast to their original assumption, \emph{one Pol may correspond to multiple instances} (i.e., $n$ instances where $n=9$ in RetinaNet~\cite{lin2017focal} and YOLO~\cite{redmon2018yolov3}). it is a more intuitive hypothesis due to the common aggregations and occlusions in the real-world. If we switch to this more general assumption, these methods would meet the computational bottleneck. Because the assumption requires them to have around $n$ times larger mask feature representations by employing a $n$ times larger standard convolutions for predicting masks, which is empirically infeasible. A smaller $n$ or a relatively low-dimensional mask presentation could avoid this issue, but their performance is compromised as mentioned. 

To address the computational feasibility in the assumption that each PoI feature may cover multiple instances, we study to generate \emph{high-dimensional instance-aware} mask representations from PoI features by our proposed \emph{instance-aware convolution}. It decomposes the mask learning into two tractable modules: \emph{instance-aware weights} and \emph{instance-agnostic features} generation (as illustrated in the bottom of Fig.~\ref{fig:sketch}). Specifically, the former module \emph{dynamically} generates $n$ convolution weights, predicting $n$ possible mask features. These dynamic weights are named as \emph{instance-aware weights} as they correlate with the predicted instances. Also, the latter module learns a high-dimensional instance-agnostic feature, serving as a template for potential mask representations from a PoI feature. By convolving the instance-agnostic feature with the instance-aware weights as convolution parameters, we will get the instance-aware feature, aligning the template feature to the specific instance. 
In this way, it is easy to focus on predicting masks for those positive instance candidates of PoIs while maintaining a large mask feature size. Thus, our proposed instance-aware convolution can solve the computational burden existing in the standard convolution for PoI-based methods.

Along with the instance-aware convolution, we propose PointINS, a simple and practical approach to evolve one-stage detectors for instance segmentation. It provides compelling PoI features for final mask prediction both efficiently and effectively.

The contributions of this study are three-fold:
\begin{itemize}
	\item We propose PointINS, a new pointed-based framework for instance segmentation equipped with the given instance-aware convolution. This module addresses the learning problem about computing proper mask representations for several instances with PoI features, concerning both effectiveness and efficiency.
	
	\item Our proposed method gives the state-of-the-art instance segmentation performance among existing PoI-based methods. This performance is also competitive compared with the region-based Mask R-CNN. Under the single-scale 1x training schedule, we obtained \textbf{34.5} mask mAP with ResNet101-FPN. Using data augmentation or with longer training time improved performance consistently to \textbf{38.3}, which is comparable to Mask R-CNN (38.3) yet with faster inference speed (14.9 fps vs. 13.5 fps (frame per second)).
	
	\item Our given PointINS is compatible with the most dense one-stage detectors (RetinaNet \cite{lin2017focal} and FCOS \cite{tian2019fcos}), as its core module instance-aware convolution is decoupled from box heads. It serves as a general framework for point-based instance segmentation.
	
\end{itemize}
\section{Related Work}
In this section, we first revisit representative studies on object detection, as its close relationship to instance segmentation. Then, we describe methods related to instance segmentation and the core technique of dynamic weights employed in this study.
Wrapped with dynamic weights, the proposed instance-aware convolution could evolve the dense one-stage detectors for instance segmentation with minimal modifications.

\subsection{Object Detection}
Current convolutional neural network (CNN)-based object detectors can be categorized into two-stage detectors or one-stage detectors based on whether they exploit region proposals. Two-stage detectors, such as a series of R-CNN~\cite{girshick2014rich,DBLP:conf/iccv/Girshick15,DBLP:conf/nips/RenHGS15} first select RoIs and then extract the region features to localize and classify objects. The following algorithms are proposed to improve the performance of two-stage detectors, including new architecture designs, context and attention mechanism, multi-scale exploration, training strategy, and loss design, feature fusion and enhancement, better proposal, and balance. 
We note that all these methods use the region feature extraction to detect objects. Due to the accurate alignment between region locations and features, RoI Align~\cite{DBLP:conf/iccv/HeGDG17} has become the primary region feature extraction method in two-stage detectors.

\begin{figure}[t]
\begin{center}
\includegraphics[height=1.2in, width=\linewidth]{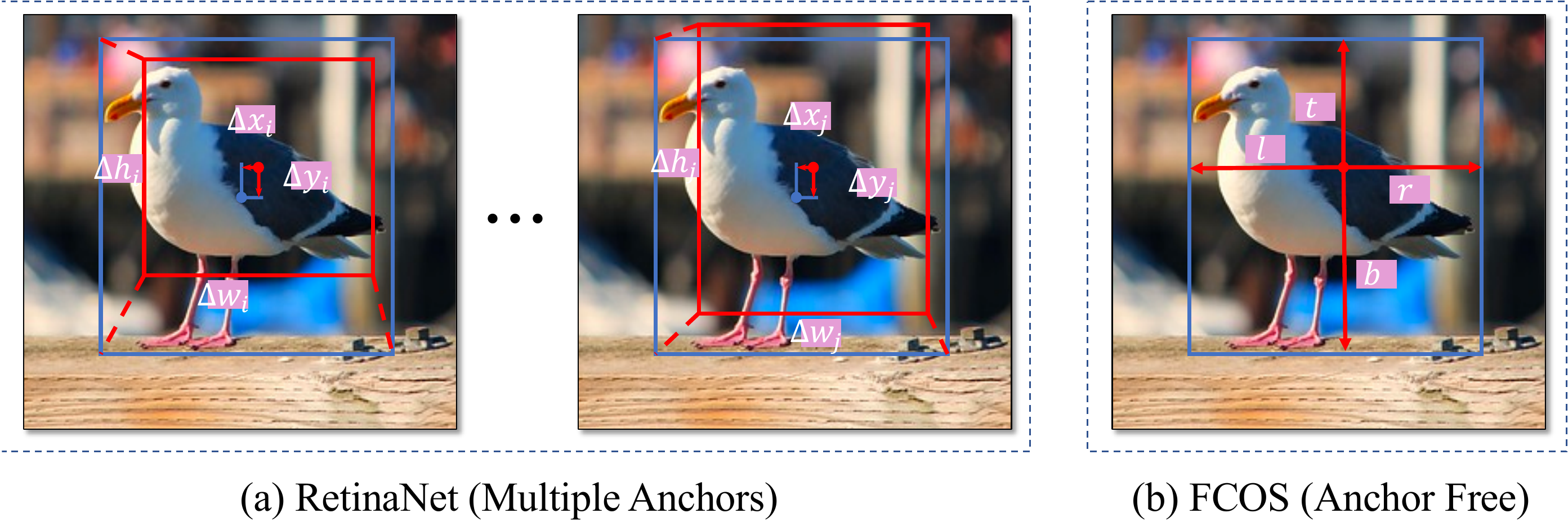}
\end{center}
\caption{Illustration of differences between RetinaNet~\cite{lin2017focal} and FCOS~\cite{tian2019fcos}. Blue points and boxes represent the center and bound of an object, respectively, while red points and boxes represent the center and bound of anchor, respectively. (a) The single point in RetinaNet responds to several predefined anchors with various sizes and aspect ratios. Thus, RetinaNet regresses from the anchor box with four offsets. (b) FCOS has no anchors, and regresses from a center point with four distances. For simplicity, we refer to this center point as an anchor point.}
\label{fig:detectors}
\end{figure}

Instead of using cropped region features, one-stage methods~\cite{liu2016ssd,redmon2016you,lin2017focal,tian2019fcos} detect boxes only from point features. Owing to the focal loss~\cite{lin2017focal}, one-stage methods can effectively balance the training loss between positive and negative samples, thereby leading to satisfactory detection performance. With faster inference speed, one-stage detectors have become increasingly popular in the computer vision area. 
Among numerous one-stage detectors, some detectors~\cite{liu2016ssd,redmon2016you,lin2017focal,tian2019fcos} generate bounding boxes from the single center point, as shown in Fig.~\ref{fig:detectors}. For simplicity, we refer to these center-point based one-stage detectors as \emph{dense one-stage detectors.} 

Like the box generation of dense one-stage detectors, our proposed PointINS framework generates instance masks from a single-point feature. Because both boxes and masks are different representations of instances, our method can be built upon these detectors regardless of whether they are anchor-based or anchor-free. This means that dense one-stage detectors can be quickly converted into instance segmentation frameworks by our method with minor modification. 

\subsection{Instance Segmentation}
Current instance segmentation methods can be roughly classified into two categories: segmentation-based~\cite{liu2017sgn,bai2017deep,wang2019solo} and detection-based~\cite{DBLP:conf/iccv/HeGDG17,DBLP:journals/corr/abs-1803-01534,chen2019hybrid,huang2019mask,lee2020centermask,peng2020snake} methods. 
Segmentation-based methods usually first predict a semantic map and then cluster points with similar semantic feature embedding into instance masks. Despite the presented new problem formulation of instance segmentation, their expensive clustering produces prevent their real-world applications.
Other types of methods encode the instance information into each channel of the generated semantic map. InstanceFCN~\cite{dai2016instance-fcn} and FCIS~\cite{li2016fully} propose predicting position-sensitive score maps with vanilla fully convolutional networks (FCNs). SOLO~\cite{wang2019solo} directly decodes each channel of the semantic map into instance masks without indicating whether the used regions are valid. The efficiency of these methods is a concern, as they segment numerous background areas containing no instances but usually covering a large portion of the images. 

Popular instance segmentation frameworks are mainly detection-based~\cite{girshick2014rich,DBLP:conf/iccv/Girshick15,DBLP:conf/nips/RenHGS15,DBLP:conf/nips/DaiLHS16,DBLP:conf/cvpr/LinDGHHB17,peng2020snake}. They generally predict a series of bounding boxes by two-stage detectors and then segment masks. Mask R-CNN is an representation for two-stage instance segmentation framework. Several methods, including PANet~\cite{DBLP:journals/corr/abs-1803-01534}, HTC~\cite{chen2019hybrid}, and Mask Scoring R-CNN~\cite{huang2019mask}, refine structure details from different aspects, and all methods use the RoI-Align~\cite{DBLP:conf/iccv/HeGDG17} to produce region features. Even RetinaMask~\cite{fu2019retinamask}, Centermask~\cite{lee2020centermask} and Embedmask~\cite{ying2019embedmask}, which are built upon one-stage detectors, such as RetinaNet~\cite{lin2017focal} or FCOS~\cite{tian2019fcos}, use RoI features to segment instances.

Instead of using RoI features, Tensormask~\cite{chen2019tensormask}, Polarmask~\cite{xie2020polarmask}, and MEInst~\cite{zhang2020MEInst} force a single-point feature to segment only one or two potential regions for exploring a more straightforward representation of instance masks. 
YOLACT~\cite{bolya-iccv2019} assembles the point responses to several prototypes for final mask prediction. However, the prototypes are predefined and involve additional hyper-parameters. In contrast to these methods, our framework generates a point-wise instance-agnostic feature, serving as a template for potential instance masks of generated bounding boxes. 

\noindent \textbf{Single-shot instance segmentation}
Single-shot instance segmentation has two popular definitions in the literature. The first depends on whether the proposed method directly generates masks without a box indicator, while the second depends on whether the proposed methods have fewer information processing steps than the popular Mask R-CNN.

SOLO~\cite{wang2019solo} and Polarmask~\cite{xie2020polarmask} belong to the former category, while Tensormask~\cite{chen2019tensormask} and MEInst~\cite{zhang2020MEInst} belong to the latter category. Without boxes, SOLO and Polarmask have to segment many invalid regions. Although Tensormask~\cite{chen2019tensormask} and MEInst~\cite{zhang2020MEInst} simultaneously generate boxes and masks, they still require box information to interpolate masks from fixed sizes to the box size, where the setting is the same as ours. 
The main difference between these methods and ours is our adopted sampling strategy, as described in subsection~\ref{sub:cc}. We sample valid regions to segment masks in advance, following the process of Mask R-CNN: detection first, followed by instance segmentation. Our premise is that focusing on valid regions can reduce the computational cost of mask generation. 

\subsection{Dynamic Weight}
As a specific strategy of meta-learning~\cite{andrychowicz2016learning,ravi2016optimization,wang2016learning}, weight prediction~\cite{lemke2015metalearning,yang2019condconv} aims to generate dynamic parameters according to different inputs. In other words, the network parameters are input-dependent. It can significantly improve the flexibility of the used net structure at the cost of additional computation.

Cai et al.~\cite{cai2018memory} used the predicted parameters of a classifier to discriminate new categories. These parameters were generated by the memory of the support set. In object detection, MetaAnchor~\cite{yang2018metaanchor} is a flexible anchor generator to produce higher quality boxes. The weight of the last regression layer is predicted with properties of customized prior anchors. To balance the instance annotation cost between box and mask, Hu et al.~\cite{hu2018learning} proposed a semi-supervised method to segment zero-shot instances by re-weighting box features. Besides, MetaSR~\cite{hu2019meta} uses a dynamic up-sampling module to super-resolve a single image with arbitrary scale factors.

In our proposed method, we use dynamic convolution weights to learn different instance representations from a single point feature, achieving alignment between an instance-agnostic feature and positive instances. With instance properties, we generate several unique instance-aware features for mask prediction while maintaining the high mask representation capacity. In this way, our method is effective and robust to one-stage dense detectors even with numerous anchors tiled in a single point.

\section{Our Method}

\begin{figure*}[ht!]
	\begin{center}
		\includegraphics[width=1\linewidth]{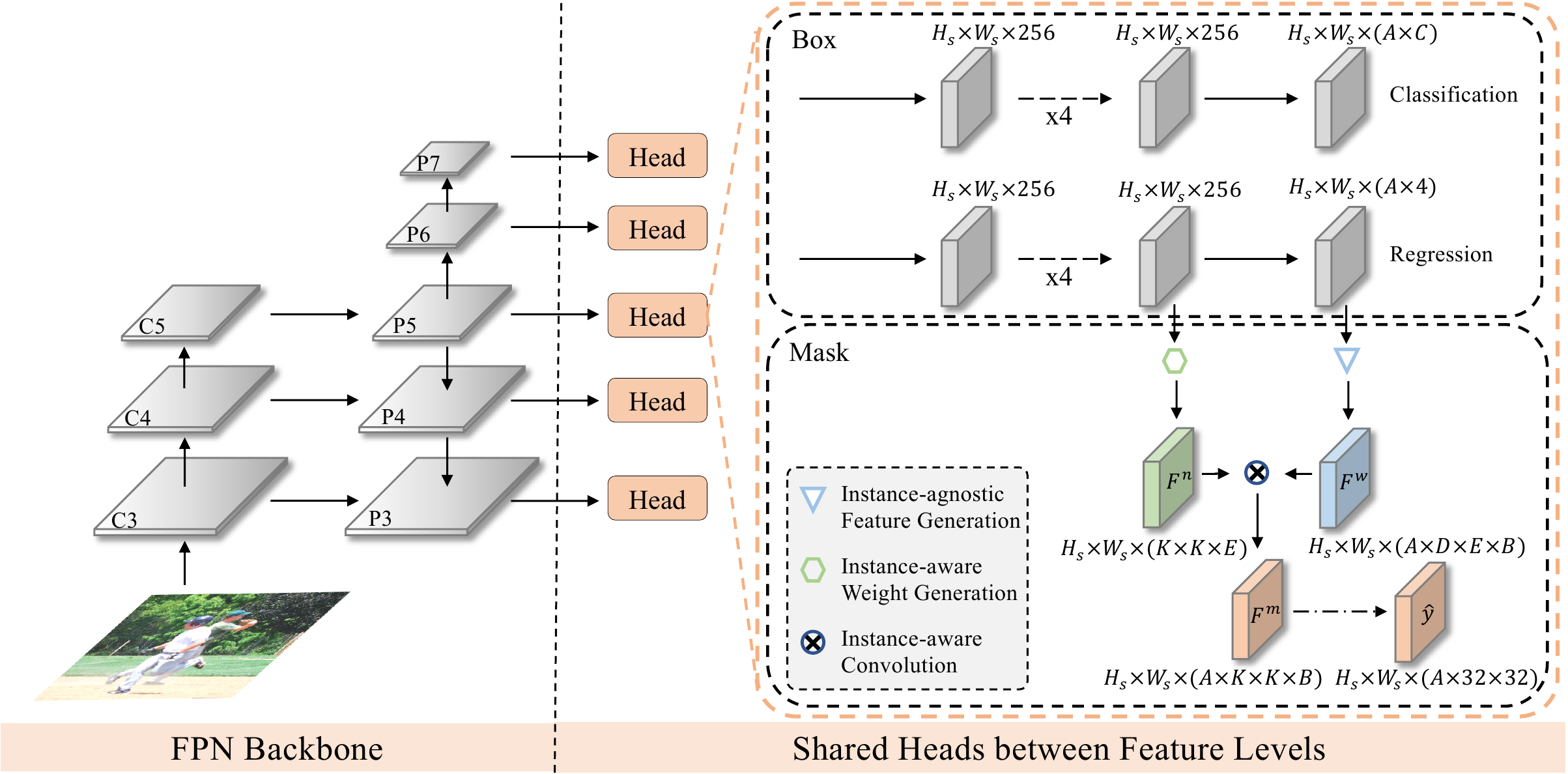}
	\end{center}
	\caption{overall architecture of proposed PointINS framework.  $C_3$, $C_4$ and $C_5$ are the feature maps of the backbone network (\textit{e.g.}, ResNet50). $P_3$ to $P_7$ are the feature pyramid network (FPN) feature maps, as in \cite{DBLP:conf/cvpr/LinDGHHB17,lin2017focal,tian2019fcos}. For better illustration, each feature map is represented by three dimensions, including height H, width W and channel without considering the image batch. A is the number of anchors covered by a single point, with $9$ and $1$ in RetinaNet and FCOS respectively. C is the number of classes, such as 80 in COCO~\cite{lin2014microsoft} dataset.}
	\label{fig:architecture}
\end{figure*}

Instance segmentation aims to detect all instances from images at the pixel level, predicting the location of an object and the category that it belongs to.
In this study, the proposed framework follows the design philosophy of Mask R-CNN, which splits the process of determining the class and location of objects into two independent sub-tasks: instance location regression and class prediction. In Mask R-CNN, both sub-tasks are predicted with RoI features extracted by RoI Align.
In contrast, this paper studies how to exploit PoI features to learn mask representation for instances. Meanwhile, the classification component follows previous point-based detectors, such as RetinaNet~\cite{lin2017focal} or FCOS~\cite{tian2019fcos}.

Generally, our proposed PointINS framework has two components in its structure: a one-stage detector and instance-aware mask prediction. We use the dense one-stage detector here, as it naturally offers pixel-level semantics via classification and our desired PoI features for mask estimation. More importantly, it generally runs faster than a two-stage detector. The employed dense one-stage detector has a classical structure, including a backbone (\textit{e.g.} ResNet~\cite{DBLP:conf/cvpr/HeZRS16}), a feature pyramid structure (\textit{e.g.} FPN~\cite{DBLP:conf/cvpr/LinDGHHB17}), and a box generation head. Our proposed instance-aware mask prediction module provides simple and effective mask estimation and is compatible with most dense one-stage detectors whatever they are anchor-based or anchor-free. It is plugged into the box generation head of the used detectors, transforming PoI features into instance masks with minor modifications.

In the remainder of this section, we first briefly revisit the applied one-stage detectors~\cite{lin2017focal,tian2019fcos}. Then, we describe our proposed instance-aware mask prediction module and demonstrate its flexibility for both RetinaNet~\cite{lin2017focal} and FCOS~\cite{tian2019fcos} with two variants. Thereafter, we analyze the advantage of our proposed module in terms of computation cost, and introduce the final loss function for joint training with the utilized one-stage detectors. Finally, we discuss the relationship between our modules and some existing works, e.g., Tensormask~\cite{chen2019tensormask}.

\subsection{Dense One-stage Detectors}
\label{subsec:detectors}
The employed detector performs the pixel-level class prediction task and provides PoI features for mask prediction.
Structurally, as illustrated in Fig~\ref{fig:architecture} (not including the mask component), it contains an FPN backbone and a detection head, utilizing anchors (predefined sliding windows) for distinguishing possible multiple instances of varied forms. 

The FPN backbone adopts a feature pyramid scheme to detect objects with different sizes. Specifically, the FPN backbone extracts feature maps $F_s \in \mathcal{R}^{H_s \times W_s \times 256}$ of different resolutions from the input image $I \in \mathcal{R}^{H \times W \times 3}$. Here, $H$, $W$, $H_s$, $W_s$ denote the height and width of an image and generated pyramid feature maps, respectively, where $s \in \{8, 16, 32, 64, 128\}$ is the stride of the pyramid feature maps compared to the input image size. 

The detection head has two branches: the classification branch and regression branch. Each branch has four convolutional blocks, including the convolutional layer and rectified linear unit (ReLU) operation. These blocks are shared among all feature-level maps. The classification branch provides the classification prediction for every pixel of the feature map with a multi-class binary vector~\cite{lin2017focal} form as $P_{s,i,a} \in \{0, 1\}^{C} $ where $P_s \in \mathcal{R}^{H_s \times W_s \times A \times C}$. $A$ and $C$ denote the anchor and object category number, respectively; and $i$ and $a$ denote the pixel location and anchor index, respectively. Also, the regression branch predicts the offset $R_s \in \mathcal{R}^{H_s \times W_s \times A \times 4}$ between the anchors and ground truth. We note that this offset regression is different when using different dense one-stage detectors. As illustrated in Fig.~\ref{fig:detectors}, RetinaNet~\cite{lin2017focal} tiles multiple preset anchors for each point of the feature map. Thus, a single-point feature is used to regress the distances between anchors and nearest objects. FCOS~\cite{tian2019fcos} does not use anchors and uses a single-point feature to learn the distances between the coordinate of points and the nearest object. A point in FCOS is similar to an anchor without restrictions on the area size and aspect ratios. From this perspective, we can regard FCOS as an anchor-based detector with only one anchor. Unlike RetinaNet, FCOS has an additional centerness branch in box generation head to predict whether the anchor point is in the center of the ground truth bounding boxes.

In our method, we regard both the bounding boxes and pixel-level masks as representations of the instances' location. Thus, our proposed module utilizes the last feature map $F^l_s$ in the regression branch for further instance segmentation.

\begin{figure*}[t!]
	\begin{center}
		\includegraphics[height=2.0in, width=1\linewidth]{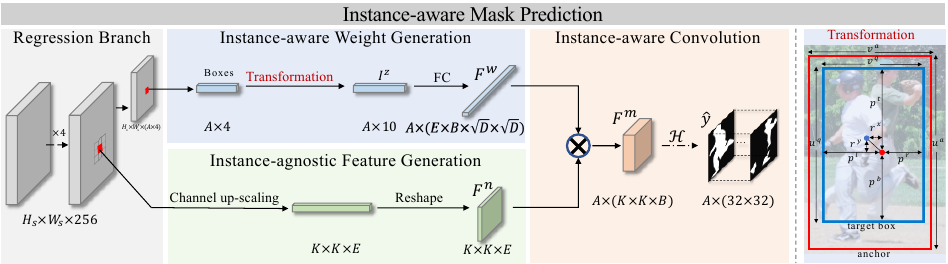}
	\end{center}
	\caption{Detailed structure of proposed instance-aware mask prediction module. For simplicity, we only sample a single-point feature generated from the $3\times3$ points feature in regression branch for illustration. The key process in proposed module is instance-aware convolution, including the \emph{instance-agnostic} feature and \emph{instance-aware} weight. First, we generate the instance-agnostic feature $F^n$ by a channel up-scaling convolution layer and a reshape operation. Meanwhile, the instance-aware weight $F^w$ is obtained by transforming the explicit instance information $I^z$. Then the two features serve as the input and weight of the \emph{instance-aware} convolution, in order to produce instance-aware feature $F^m$ for final mask prediction. The right-most part is a geometric illustration of our transformation module.}
	\label{fig:framework}
\end{figure*}
\subsection{Instance-aware mask prediction}
For our proposed instance-aware mask prediction, each PoI feature $F^l_{s,i} \in \mathcal{R}^{1 \times 1 \times 256}$ is extracted from the regression branch of the one-stage detector. It produces a $K \times K$ class-agnostic binary mask for every anchor as $M_{s,i,a} \in \mathcal{R}^{K \times K} $, where $M_s \in \mathcal{R}^{H_s \times W_s \times A \times K \times K}$ is the regression result from the mask feature $F^m_s \in \mathcal{R}^{H_s \times W_s \times A \times K \times K \times B}$. $K \times K$ and $B$ denote the mask resolution and mask feature dimension, respectively. 

A main advantage of the proposed instance-aware mask prediction is that it supports a high-dimensional mask feature with sufficient characteristics from PoI features, thereby exhibiting the comparable instance segmentation performance for those methods using RoI features. For example, the mask feature size $K^2B$ in Mask R-CNN is $50176=256\times14^2$. Our support for high-dimensional mask representation in PoI form results from our decoupled design of the mask feature computation. 
For each point, we decouple this computation into instance-agnostic features and instance-aware weights, in which we project the shared template to the unique mask feature. 
Specifically, with the feature maps from the regression branch of the dense detectors, we disentangle the mask feature computation into two parts: instance-agnostic features $F_s^n \in \mathcal{R}^{H_s \times W_s \times K^{2}E}$ and instance-aware weights $F^w_s \in \mathcal{R}^{H_s \times W_s \times A\times (D\times E) \times B}$. $D$ is the weight kernel size, which is described in Section~\ref{sec:implementaion}.

As the name suggests, every instance-agnostic feature point $F^n_{s,i} \in \mathcal{R}^{K \times K \times E}$ indicates a primitive template feature containing robust information on its receptive field.
Thus, this feature can be responsible for all tiled anchors.
Besides, every instance-aware weight point $F^w_{s,i} \in \mathcal{R}^{A \times B \times D}$ represents $A$ mask feature embeddings whose embedding dimension is $B\times D$, derived from the anchors' property (such as anchor width/height or aspect ratio) and the regression offset between anchors and target boxes. 
Using dynamic convolution as the feature alignment, we transform the shared template feature to the unique feature of the potential instance.
\begin{equation}
F^m_{s,i} = \bigotimes(F^n_{s,i}, F^w_{s,i}),
\end{equation}
Where $\bigotimes$ is the dynamic convolution, in which $F^n_{s,i}$ is input and $F^w_{s,i}$ is the dynamic convolution weight.

In structure, the instance-agnostic features and instance-aware weights are extracted from two different branches; however, both stem from the regression branch of the one-stage detector. They are illustrated in Fig.~\ref{fig:framework}, and their specific designs are discussed below.

\subsubsection{Instance-agnostic Feature Generation} This module aims to provide robust and informative template features for the numerous anchors. In our framework, a single-point feature $F_{s,i}^n$ is capable of producing several masks. Thus, the single-point feature should have a sufficiently large receptive field to cover all possible masks, and contain sufficient semantic information for subsequent mask prediction. This conjecture was verified in our ablation study, as illustrated in Table~\ref{Tab:template}.

Let $F_{s,i}^n$ denote the transformed features from the last feature map in the regression branch of one-stage detectors. It is computed by an additional \emph{channel up-scaling convolution} and a reshape operation to further aggregate location information. As such, our point feature is robust enough to capture the entire instances, even if this point is in the instance boundary.

Specifically, the aforementioned channel up-scaling operation is a convolution layer with input and output channel numbers as $K^2$ and $K^2E$, respectively. The convolution kernel size is $\sqrt E \times \sqrt E$. 
Since $K^2E=\sqrt E \times \sqrt E \times K^2$, the output channel increases $E$ times with its $\sqrt E \times \sqrt E$ convolution kernel size. This operation is essential in practice, as the increased channels allow our tensor to encode more context information for mask prediction. We note that this operation can become relatively computationally cheap, as discussed in Section~\ref{sub:cc}. On the other hand, the reshape operation is used to balance the template spatial resolution (spatial dimension) and feature representation (channel dimension). We transform a single-point feature into a three-dimensional tensor with size $(K\times K \times E)$. In this way, we can force each channel of our point feature to have explicit relative-location information\footnote{It is similar to R-FCN \cite{DBLP:conf/nips/DaiLHS16} and FCIS \cite{li2016fully}, whose channels explicitly encode relative location scores of instances.}. For example, the first nine dimensions of features before the reshape operation respond to the information of the 
upper-left-most location in the template feature. 

Because the number of channels of the last feature map in the regression branch we used is 256, we set $K$ to $16=\sqrt{256}$. The $E$ is $9$ by default due to the prevalence of the $3\times 3 $ convolution layer in neural network design.

\subsubsection{Instance-aware Weight Generation}  
\label{subsec_meta} 

This module targets to generate $n$ dynamic convolution weights from a PoI feature, used for differentiating $n$ possible instances. Its dynamics prevents from employing $n$ different conventional convolutions. From this perspective, it notably reduces the used parameters and their corresponding footprints in the computation when $n$ is relatively large, e.g., $n=9$.

For its design, let $I^z_{s,i,a}$ denote the PoI feature with instance information, we predict their corresponding convolution weight \(F^w_{s,i,a}\) by an FC layer, followed by a ReLU operation as: 
\begin{equation}
F_{s,i,a}^w = ReLU(FC(I_{s,i,a}^z)),
\label{eq:proposal_ind}
\end{equation}
where the FC layer with input channel as $U$ (10 for RetinaNet and 7 for FCOS) and output channel as $B\times D$. \(I^z_{s,i,a}= \{I^o_{s,i,a}, I^q_{s,i,a}\}\) can be divided into an anchor indicator \(I^o_{s,i,a}\) and an instance indicator \(I^q_{s,i,a}\). \(I^o_{s,i,a}\) contains information about which specific anchor at a specific feature level corresponds to the potential mask. \(I^v_{s,i,a}\) expresses the detail offsets between an anchor and its regressed instance. These two indicators are detailed below.

For \(I^o_{s,i,a}\), it has four components as follows:
\begin{equation}
I^o_{s,i,a} = \{\frac{1}{s}, \frac{h_{s,i,a}}{w_{s,i,a}}, \frac{h_{s,i,a}}{s}, \frac{w_{s,i,a}}{s}\},
\label{eq:I_A}
\end{equation}
where \(s\) is the stride of the global feature level, and \(h_{s,i,a}\) and \(w_{s,i,a}\) are the height and width of the anchors, respectively. Therefore, the last three components represent the aspect ratio and scale of the anchor.

Moreover, \(I^q_{s,i,a}\) represents the instance offset from the original anchors as
\begin{equation}
I^q_{s,i,a} = \{r^{x}_{s,i,a}, r^{y}_{s,i,a}, p_{s,i,a}^l, p_{s,i,a}^r, p_{s,i,a}^b, p_{s,i,a}^t\},
\label{eq:proposal_ind}
\end{equation}
where \(r^x_{s,i,a}\) and \(r^y_{s,i,a}\) represent the distances between the centers of the anchor and target box, respectively, and the other elements represent the distances between the box location and the center of the corresponding anchor in four directions. Each component in \(I^q_{s,i,a}\) is calculated as
\begin{equation}
\begin{split}
r^{x}_{s,i,a}=\frac{{x}^{q}_{s,i,a}-{x}^{o}_{s,i,a}}{s}, \ \ \ & r^{y}_{s,i,a}=\frac{{y}^{q}_{s,i,a}-{y}^{o}_{s,i,a}}{s} \\
p_{s,i,a}^l = \frac{0.5*w_{j}^q}{s}-r^{x}_{s,i,a}, \ \ \ & p_{s,i,a}^r = \frac{0.5*w_{j}^q}{s}+r^{x}_{s,i,a} \\
p_{s,i,a}^t = \frac{0.5*h_{j}^q}{s}-r^{y}_{s,i,a}, \ \ \ & p_{s,i,a}^b = \frac{0.5*h_{j}^q}{s}+r^{y}_{s,i,a} \\
\end{split}
\label{eq:proposal_detail}
\end{equation}
where \(x^q_{s,i,a}\) and \(x^o_{s,i,a}\) are the x-axis center location of target box and anchors, respectively, and are the same as \(y^q_{s,i,a}\) and \(y^o_{s,i,a}\). \(h_{s,i,a}^q\) and \(w_{s,i,a}^q\) are the height and width, respectively, of the target box. The rightmost part of Fig.~\ref{fig:framework} presents the geometric illustration of our transformation module.

Note that \(I_{s,i,a}^z\) only contains \(I_{s,i,a}^q\) and $\frac{1}{s}$ for the anchor-free detectors. Using FCOS~\cite{tian2019fcos} as an example, we regard the points of each feature level as our anchors, in which $x^o_{sia}$ and $y^o_{sia}$ are the sampled-point locations of the images.

\begin{figure*}[t]
	\begin{center}
		\includegraphics[width=0.9\linewidth]{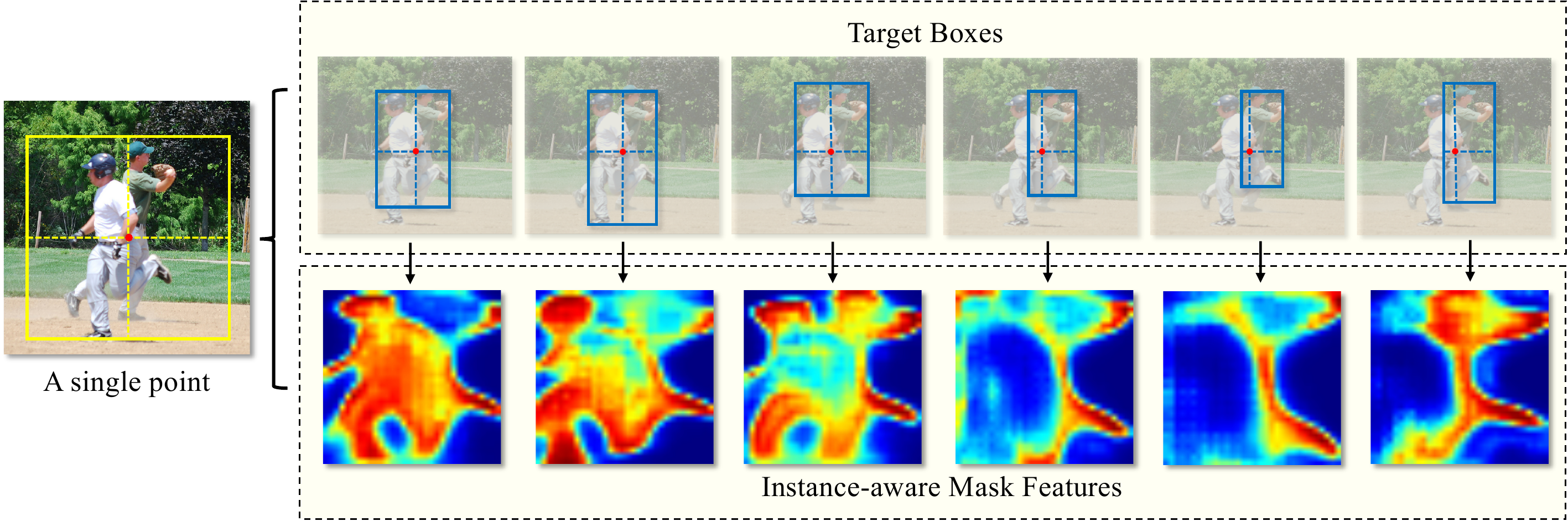}
	\end{center}
	\caption{Visualization of instance-aware features under the same point feature. With various instance information, instance-aware convolution generates distinct instance-aware features for subsequent mask prediction.}
	\label{fig:vis_feature}
\end{figure*}

\noindent \textbf{Effectiveness of instance-aware Weights} Fig.~\ref{fig:vis_feature} presents a visualization of an instance-aware feature by using different instance-aware weights for the same instance-agnostic feature. The dynamic weights can produce various instance-correlated features from the shared instance-agnostic feature.

\subsubsection{Instance-aware convolution}
\label{sec:implementaion}
The instance-agnostic feature and instance-aware weights are fused by pixel-wise convolution, leading to the mask features. We reshape the instance-aware weight $F_{s,i,a}^w$ with the $E\times B\times D$ dimension to the $E \times B \times \sqrt D \times \sqrt D$ dimensions.
The reshaped weights can be regarded as convolution weight with input channel $E$, output channel $B$, and kernel size $\sqrt D$.
For $a$th anchor of the point in location $i$, we obtain our instance-aware mask feature $F_{s,i,a}^m$ by the dynamic convolution as: 
\begin{equation}
F_{s,i,a}^m = {\bigotimes}(F_{s,i}^n , F_{s,i,a}^w).
\label{eq:RoI}
\end{equation}
where the input of this convolution is the instance agnostic template $F_{s,i}^n$, and it is parametrized by the instance-aware weights $F_{s,i,a}^w$.

In this way, our instance-agnostic feature can be aligned to the corresponding mask feature with $K \times K \times B$ dimension. Then via the standard mask head of Mask R-CNN including four convolution blocks and a deconvolution block, it is prepared for final class-agnostic mask prediction. 
Defining the mask head operation as $\mathcal{H}$, we get the final mask prediction as: 
\begin{equation}
\hat{y}_{s,i,a} = \mathcal{H}(F_{s,i,a}^m)
\label{eq:y}
\end{equation}

\subsection{Computational Efficiency Optimization}
\label{sub:cc}
In our proposed method, instance masks are only generated from a PoI feature. This generation method allows our approach to be easily improved for high efficiency. We can realize this by only sampling positive points and generating their corresponding instance-agnostic feature and instance-aware weight in parallel.

At first, a point is defined as positive if the Intersection of Union (IoU) between the ground truth and one of the anchors and target boxes is larger than $0.5$. In the COCO dataset, there are $6.43$ positive points for each ground truth instance following this sample principle with only considering the anchor condition. We randomly sample the maximum 128 points in training out of more than $22,000+$ points on all feature levels. For inference, only 100 detected points are used for final mask prediction. Note that the same point is likely to be sampled multiple times due to several of its covered anchors and target boxes satisfying our sample principle.

Thus, the subsequent processing is applied only on these points, which enables remarkable acceleration.
Due to the channel up-scaling convolution with kernel size 3 in the instance-agnostic feature generation module, it is only necessary to sample $3\times3$ receptive fields from the feature map to produce our instance-agnostic features. 
For the instance-aware weights, it is only necessary to sample the corresponding four-dimensional regression feature for each point.
Also, because these points are independent, they can be processed in parallel for further notable acceleration and memory usage reduction. Our inference achieves 15.8 FPS with ResNet50 backbone shown in Table~\ref{Tab:FPS}.

\subsection{Loss function}
The overall optimization goal considers both instance mask segmentation and the employed detection as
\begin{equation}
\mathcal{L} = \lambda_{det} \cdot \mathcal{L}_{det} + \lambda_{mask} \cdot \mathcal{L}_{mask},
\label{eq:L}
\end{equation}
where $\mathcal{L}_{det}$ and $\mathcal{L}_{mask}$ denote the detection loss and mask loss, respectively. The balance weights in $\lambda_{det}$ and $\lambda_{mask}$ are set to 1 and 2. 

For $\mathcal{L}_{det}$, it relies on the one-stage detector used. For example, RetinaNet consists of $\mathcal{L}_{cls}$ for classification and $\mathcal{L}_{reg}$ for bounding box regression. FCOS also has an extra loss $\mathcal{L}_{cen}$ for center-ness. We adopt the default setting for these detection losses. $\mathcal{L}_{cls}$ and $\mathcal{L}_{cen}$ are the focal loss and binary cross-entropy loss, respectively. $\mathcal{L}_{reg}$ is the smooth-l1 loss in RetinaNet or the generalized IoU loss in FCOS. 

For $\mathcal{L}_{mask}$, it is defined as
\begin{equation}
\mathcal{L}_{mask} = \sum_{s}^{S} \sum_{i}^{H\times W}\sum_{a}^{A}\mathbbm{1}^{obj}d_{mask}(\hat{y}_{s,i,a},y_{s,i,a})
\label{eq:L_MASK}
\end{equation}
where $\mathbbm{1}$ is the indicator function for positive samples. $\hat{y}$ and $y$ denote the prediction and ground-truth vectors, respectively. The $d_{mask}$ is the weighted binary cross-entropy loss. The edge of the mask is emphasized by assigning the larger weight as 4.

Finally, our proposed module is jointly trained with the detectors it is built upon. 

\subsection{Relation to Existing Methods}
As the PoI-based methods, e.g., Tensormask~\cite{chen2019tensormask}, we adopt PoI features to predict instance masks.
However, we provide a new mask representation computation design with dynamic weights, and our mask representation is more effective and efficient than the existing works~\cite{chen2019tensormask,xie2020polarmask,zhang2020MEInst} due to its high capacity, as validated in the performance comparison presented in Table~\ref{Tab:overall}.

We obtain our instance-aware features by instance-aware convolution. The core of this step is the instance-aware weights derived from the anchor property and box regression feature. These weights dynamically interpolate the instance-agnostic feature to the instance-ware mask feature. Unlike bilinear interpolation used in Tensormask, dynamic convolution can flexibly align the instance-agnostic feature to a specific region by learnable weights. Thus, our module can be easily generalized to dense detectors with multiple anchors. With the sampling strategy, our module only focuses on PoIs while maintaining a high mask representation.

In a comparison regarding mask representation, our mask feature size is $65536=256\times16^2$ from PoI features with $K=16$ and $B=256$, outperforming the mask feature size (only 256) of Polarmask~\cite{xie2020polarmask} and MEInst~\cite{zhang2020MEInst}. The low-dimensional mask feature of PolarMask and MEInst results in reduced performance regardless of how the final mask representation is modeled. The outlier is Tensormask~\cite{chen2019tensormask} with a dynamic mask feature size from $15\times15$ to $480\times480$ by bilinear interpolation.  This usage of bilinear interpolation depends on the aligned representation and tensor pyramid structure they proposed; however, this leads to 
a computationally intensive feature process.

Moreover, due to the proposed instance-aware convolution, our framework is robust to the anchor number each PoI feature corresponds to, making it applicable to the most types of dense one-stage detectors (both anchor-based and anchor-free), while existing methods only work in one specific type.

\section{Experiments}

\subsection{Experiments on COCO Dataset~\cite{lin2014microsoft}}
We compare our method with other state-of-the-art approaches on the challenging COCO dataset~\cite{lin2014microsoft}. Following common practice~\cite{DBLP:conf/iccv/HeGDG17,DBLP:conf/cvpr/LinDGHHB17,DBLP:journals/corr/abs-1803-01534,huang2019mask}, we train our models with 115,000 train images and reported results on the 5000 validation images for the ablation study. We also report results on 20,000 test-dev images for comparison. 

Comprehensive ablation studies are conducted on this dataset with RetinaNet-ResNet50-based PointINS. 
We follow standard evaluation metrics, namely, average precision for IoU from $0.5$ to $0.95$ with a step size of $0.05$ ($AP$), $AP_{50}$, $AP_{75}$, $AP_{S}$, $AP_{M}$ and $AP_{L}$. The last three measure performance with respect to objects in different scales.

\textbf{Training Details}
We train our network using batch size $16$ for 12 epochs. The shorter and longer edge sizes of the images are $800$ and $1333$. Adam gradient descent with learning rate as $0.00005$ , beta as $(0.9, 0.999)$, and eps as $1e^{-08}$ is used as the optimizer. We decay the learning rate with $0.1$ at $8$ and $11$ epochs respectively. Using stochastic gradient descent leads to a performance reduction of approximately $0.3$. We initialize our backbone networks with the pre-trained models on ImageNet. Moreover, we sum all the losses directly, \textit{i.e.}, $\lambda_{det}=1,\lambda_{mask}=2$ in Eq. \ref{eq:L}.

\textbf{Inference Details}
The inference is kept the same as dense detectors, as we only append one additional mask prediction to the predicted boxes. Two strategies are used to obtain the final result on the condition of whether or not the samples have valid boxes. The former generates all the masks and then samples a maximum of 100 instances by box non-maximum suppression (NMS) following Tensormask~\cite{chen2019tensormask}. The latter uses box NMS to sample a maximum of 100 boxes and then predict their masks, like MEInst~\cite{zhang2020MEInst}. Compared to the former strategy, the latter does not lead to performance reduction but avoids the unnecessary computational overhead for masks. 

\subsubsection{Ablation Study}
All ablation studies are conducted by RetinaNet-based PointINS with ResNet50. Note that this baseline yields 32.5 AP for the validation split.

\textbf{Instance-agnostic Feature Generation}
The representation of the template tensor is first studied. As illustrated in Table \ref{Tab:template}, we explore the influence of different shapes on the performance of the instance-agnostic feature. For a fair comparison, we fix the kernel size $E$ of the channel-up scaling convolution layer to $3$. That means, we only change the output channel of this convolution layer to match the size of $(E, K, K)$ of Table~\ref{Tab:template}. The default tensor shape is $(9, 16, 16)$, where 9 is the feature representation and 16 is the spatial resolution. The performance decreased when we reduced the number of feature representations. This indicates that the feature representation dimension is vital for encoding information.

Intriguingly, as illustrated in the last two rows, further increasing the number of feature representation does not improve the results. This is resulted from our used template tensor. Regardless of how we transform the information, the template is reshaped from a single-point feature. In this ablation study, this single-point feature is generated by a $3\times 3$ convolution with an input channel number of 256 by default. Therefore, the information capacity is $3 \times 3 \times 256$. The larger output channels of convolution do not further increase the capacity of valid messages.

This analysis also explains the performance reduction using different reshape styles. A $3\times 3$ kernel
encodes the $9$ spatial locations in sequence. Therefore, reshaping to a $(9,16,16)$ tensor is more intuitive than the manner in which each channel represents the relative location response. We note that this is also the core principle of R-FCN~\cite{DBLP:conf/nips/DaiLHS16} and FCIS~\cite{li2016fully}. 
We regard a tensor with size $(1,16,16)$ as the nature representation~\cite{chen2019tensormask} proposed by the TensorMask study. It causes a reduction of approximately 3 mean average precision (mAP).

\begin{table}[t!]
    \centering
    \caption{Ablation study for instance-agnostic feature representation with a fixed kernel size of 3 of channel up-scaling convolution. The reshaped tensor' size is $(E, K, K)$, where $E$ denotes the feature representation and $K$ denotes the spatial resolution. The Modification column lists the methods of changing the tensor size and shape by modifying the channel up-scaling output channel or performing the reshape operation.}
    \small
    \begin{tabular}{c|c|ccc}
        \toprule
        $(E, K, K)$ & modification & $AP$ & ${AP}_{50}$ & ${AP}_{75}$ \\ \midrule
        $(9, 16, 16)$ & default & \textbf{32.5} & \textbf{51.9} & \textbf{33.8} \\ \midrule
        $(1, 16, 16)$ & \multirow{2}{*}{Decrease Channels} & 30.4 & 49.2 & 31.6\\ 
        $(4, 16, 16)$ & & 31.6 & 50.1 & 32.4\\ \midrule
        $(36, 8, 8)$ & \multirow{2}{*}{Reshape Styles} & 31.8 & 51.0 & 32.4 \\ 
        $(1, 48, 48)$ & & 29.6 & 48.5 & 30.1\\ \midrule
        $(16, 16, 16)$ & \multirow{2}{*}{Increase Channels} & 32.3 & 51.7 & 33.7 \\ 
        $(25, 16, 16)$ & & \textbf{32.5} & 51.8 & 33.6\\ \bottomrule
    \end{tabular}
    \label{Tab:template}
\end{table}

Table~\ref{Tab:template_2} outlines an ablation study using different kernel size $E$ of channel up-scaling convolution with an output channel number of $256\times E^2$. We note that this ablation study is different from that in Table~\ref{Tab:template}, with various kernel size \textit{e.g. 1, 3, 5 or 7}. Therefore, there is no capacity change via this convolution because its input channel number is always 256. We find it that using a large kernel size would not lead to a noticeable performance increase. We conjecture that a $3\times3$ kernel size is sufficient to help the convolution layer capture the target instance regions, because its input has a large enough receptive field, which has been already used for bounding box prediction.

\begin{table}[t!]
    \centering
    \caption{Ablation study for instance-agnostic feature representation with various kernel size of channel up-scaling convolution. The reshaped tensor' size is $(E, K, K)$, where $E$ denotes the feature representation and $K$ denotes the spatial resolution.}
    \small
    \begin{tabular}{c|ccc}
        \toprule
        $(E, K, K)$  & $AP$ & ${AP}_{50}$ & ${AP}_{75}$ \\ \midrule
        $(9, 16, 16)$ & 32.5 & 51.9 & 33.8 \\ \midrule
        $(1, 16, 16)$ & 30.4 & 46.3 & 28.7\\ \midrule
        $(25, 16, 16)$ & \textbf{32.7} & \textbf{52.0} & \textbf{33.9}\\ \midrule
        $(49, 16, 16)$ & \textbf{32.7} & 51.9 & 33.9 \\ \bottomrule
    \end{tabular}
    \label{Tab:template_2}
\end{table}

\begin{table}[t!]
\centering
\caption{Ablation study for the input to generate instance-aware weights. The inputs was split into three parts and explored. Each variable formulation is simplified by deleting its subscript. For example, $I^o$ denotes $I^o_{sia}$. $\circ$ represents ``not used" for this part, whereas $\checkmark$ represents used.}
\small
\begin{tabular}{c|c|c|ccc}
\toprule
$I^o$ & $\{r^x, r^y\}$ & $\{p^l, p^r, p^b, p^t\}$ & $AP$ & ${AP}_{50}$ & ${AP}_{75}$ \\ \midrule
$\circ$ & $\circ$ & $\circ$ & 27.9 & 46.5 & 28.5\\ \midrule
\checkmark & $\circ$ & $\circ$ & 25.5 & 44.9 & 26.1\\ \midrule
$\circ$ & \checkmark & $\circ$ & 26.3 & 45.0 & 26.6\\ \midrule
$\circ$ & $\circ$ & \checkmark & 28.9 & 47.1 & 29.3\\ \midrule
\checkmark & \checkmark & $\circ$ & 30.4 & 49.1 & 31.2\\ \midrule
$\circ$ & \checkmark & \checkmark & 31.7 & 50.6 & 32.8\\ \midrule
\checkmark & $\circ$ & \checkmark & 31.3 & 50.5 & 32.6\\ \midrule
\checkmark & \checkmark & \checkmark & \textbf{32.5} & \textbf{51.9} & \textbf{33.8} \\ \bottomrule
\end{tabular}
\label{Tab:meta-input}
\end{table}

\begin{table}[t!]
\centering
\caption{Performance comparison with other heuristic designs for $I_z$. Each variable formulation is simplified as in Table~\ref{Tab:meta-input}. The first column describes the design method.}
\small
\begin{tabular}{c|ccc}
\toprule
the heuristic $I^z$ design & $AP$ & ${AP}_{50}$ & ${AP}_{75}$ \\ \midrule
Generating $I^z$ from the feature map& 27.9 & 46.5 & 28.5\\ \midrule
$I^q$ has FPN's box regression style & 31.8 & 50.3 & 32.7\\ \midrule
Log$(I^o)$ & 32.2 & 51.6 & 33.6\\ \midrule
Sigmoid$(I^z)$ & 32.3 & 51.6 & 33.6\\ \midrule
Tanh$(I^z)$ & 32.3 & 51.8 & 33.4\\ \midrule
default Design & \textbf{32.5} & \textbf{51.9} & \textbf{33.8} \\ \bottomrule
\end{tabular}
\label{Tab:meta-input-2}
\end{table}

\textbf{Instance-aware Weight Generation}
Table~\ref{Tab:meta-input} summarizes an ablation study on the input used for generating instance-aware weight. We first explore the two extreme cases, presented in the first and last rows of Table~\ref{Tab:meta-input}. Our framework achieves the best performance when involving all three parts. In the first row, we directly use the features from the detection branch instead of our explicit box information. The performance reduction indicates that the explicit information of bounding boxes is more helpful in generating instance-aware weights than directly using the features. We suppose that the features directly generated by the network are insensitive to the stride of feature maps.

As alternatives to using this piece of information, only using the $I^o$ or $\{r^{x}, r^{y}\}$ leads to lower AP than directly using the features of the detection head. This means that the offset between anchors and boxes is vitally important in our module. Our design makes it possible to determine the exact shift to guide fine-tuning the instance-agnostic template. This assumption is verified in Fig.~\ref{fig:vis_feature}. In our module, $\{p^l, p^r, p^b, p^t\}$ play vital roles, and $I^o$ (and $\{r^{x}, r^{y}\}$) is peripheral.

Table~\ref{Tab:meta-input-2} presents the performance with other heuristic designs for $I^z$. The results indicate that the unique region indicators are critical to our performance. The design of the last row is our default setting. For other designs, regardless of how the hand-crafted features are designed, the final performance is relatively robust. We suppose this robustness is produced by our proposed weight prediction structure (FC Layer). This structure could handle the variations in our designed features and then produce instance-aware weights. The performance change among various element-wise operations, such as Log, Sigmoid, or Tanh, is marginal for $I^o$ or $I^q$. Using the FPN's box regression style in $I^q$ 
produces an AP lower by approximately $0.7$ than using FCOS's style. The main difference between these two regression styles has been illustrated in Fig.~\ref{fig:detectors}. 
The performance comparison (Table~\ref{Tab:meta-input-2}) shows that the boundary distance between anchor and target box (Fig.~\ref{fig:detectors} (b)) is more helpful to generate offset weight than width or height variance (Fig.~\ref{fig:detectors} (a)).

Table~\ref{Tab:meta-structure} reveals the influence of using different weight prediction structures. This table indicates that there are no large differences when a single FC layer is used and not used. The performance of two layers is only $0.1$ higher than a single layer. The reason is that our input only contains 10 elements. Using more layers does not greatly enhance information in this setting. Moreover, the ReLU operation guarantees non-linearity in our module.

\begin{table}[t!]
\centering
\caption{Ablation study for instance-aware weight prediction structure. FC and ReLU represent the fully connected layer and rectified linear unit, respectively. The ``+" symbol represents cascading two operations in sequence. $2 \times$ means repeating this operation module twice.}
\small
\begin{tabular}{c|ccc}
\toprule
structure & $AP$ & ${AP}_{50}$ & ${AP}_{75}$ \\ \midrule
FC+ReLU & 32.5 & 51.9 & \textbf{33.8} \\ \midrule
FC & 31.7 & 51.0 & 33.4\\\midrule
2$\times$(FC+ReLU) & \textbf{32.6} & \textbf{52.0} & 33.6 \\ \midrule
FC+ReLU+FC & 32.3 & 51.7 & 33.6\\ \bottomrule
\end{tabular}
\label{Tab:meta-structure}
\end{table}

\begin{table}[t!]
\centering
\caption{Ablation study for information fusion between the instance-agnostic template and instance-aware weight. For feature addition and concatenation, we reshaped both our instance-agnostic feature and instance-aware weight to $16\times16\times9$. }
\small
\begin{tabular}{c|ccc}        
\toprule
combination & $AP$ & ${AP}_{50}$ & ${AP}_{75}$ \\ \midrule
	addition & 31.1 & 50.9 & 33.5 \\ \midrule
concatenation & 30.9 & 50.6 & 33.6\\ \midrule
3$\times$3 convolution & \textbf{32.5} & 51.8 & 33.7\\ \midrule
1$\times$1 convolution & \textbf{32.5} & \textbf{51.9} & \textbf{33.8} \\ \bottomrule
\end{tabular}
\label{Tab:meta-connection}
\end{table}

\textbf{Instance-aware Convolution}
Table~\ref{Tab:meta-connection} summarizes an ablation study on different fusion methods of the instance-agnostic template and instance-aware weight. It can be seen that directly adding or concatenating them leads to performance reduction. The reason is that these two features come from different sources and thus have a distinct meaning. Specifically, the instance-agnostic template has robust semantic information, whereas the instance-aware weight indicates shifting information between the template and proposals. As a result, using dynamic convolution is more appropriate for fusing them by interpolating semantic information with geometric offsets. With this consideration, our design is effective in aligning instance-agnostic mask features given different box information. Increasing the convolution kernel size of $1\times1$ to $3\times3$ does not further increase performance, so we choose $1\times1$ kernel, further reducing the computation cost of our instance-aware convolution.

\textbf{Sampling Strategy}
We explore the effectiveness of the sampling strategy. As illustrated in Section~\ref{sub:cc} (Computation Optimization), we randomly sample 128 positive points by default. 
The influence about the number of positive points on the performance is initially ablated, as presented in Table ~\ref{Tab:sample_num}. The mask improves as the number of positive points increases from 32 to 128. Further increasing the number of positive points do not improve performance because of the limited number of positive points in the COCO image.
\begin{table}[t]
    \centering
    \small
    \caption{The ablation study on number of sampled positive points. "all" means we sample all the positive points once they satisfy our sample criterion.}
    \begin{tabular}{c|ccc}
        \toprule
        Number of Positive Points & $AP$ & ${AP}_{50}$ & ${AP}_{75}$ \\ \midrule
        32 & 29.6 & 48.9 & 31.8  \\ \midrule
        64 & 31.7 & 51.1 & 33.0  \\ \midrule
        128 & \textbf{32.5} & \textbf{51.9} & \textbf{33.8}  \\ \midrule
        256 & 32.4 & 51.7 & \textbf{33.8}  \\ \midrule
        all & \textbf{32.5} & \textbf{51.9} & 33.7 \\ \bottomrule
    \end{tabular}
    \label{Tab:sample_num}
\end{table}

Table~\ref{Tab:sample_iou} outlines an ablation study on the IoU threshold for positive points. The second row displays our default setting with which the IoU between the ground truth and both its anchors and regressed boxes are greater than 0.5. The performance dramatically decreases when we reduce the anchor’s IoU threshold to 0.4. The reason for this is the sampling strategy adopted in the detection branch. Only anchors whose IoU are larger than 0.5 are used for box training. This means that the other anchors could not be trained; thus, the quality of their generated boxes is not guaranteed. However, increasing the IoU threshold of 0.5 to another value like 0.6 or 0.7 reduces the mask performance due to the network underfitting with fewer positive points.
\begin{table}[t]
    \centering
    \small
    \caption{The ablation study on the sample criterion for positive points. As illustrated in section~\ref{sub:cc}, a point is defined positive if the Intersection of Union (IoU) between ground truth and one of the anchors and target boxes are larger than $0.5$.}
    \begin{tabular}{c|c|ccc}
        \toprule
        IoU(Anchor, GT) & IoU(Box, GT) & $AP$ & ${AP}_{50}$ & ${AP}_{75}$ \\ \midrule
        0.4 & 0.5 & 30.2 & 50.7 & 32.7  \\ \midrule
        0.5 & 0.5 & \textbf{32.5} & \textbf{51.9} & \textbf{33.8}  \\ \midrule
        0.5 & 0.6 & 32.0 & 51.2 & 33.1  \\ \midrule
        0.5 & 0.7 & 31.0 & 50.1 & 32.2  \\ \midrule
        0.6 & 0.6 & 31.7 & 50.9 & 32.9  \\ \bottomrule
    \end{tabular}
    \label{Tab:sample_iou}
\end{table}

Due to the heuristic design for our dynamic weight prediction input $I^z$, we can sample three types of target boxes for each point, including the anchor, ground truth, and regressed box. We ablate the influence of the sampling ratio of these three types, as illustrated in Table~\ref{Tab:sample_ratio}. The performance of sampling the anchors' corresponding ground truth for all points is slightly lower than our default setting (32.4 vs 32.5). This indicates that our framework could be trained without the regressed boxes, and box information is required only to segment valid regions in inference. The performance of only using regressed boxes is inferior to our default setting. The reason for this is that there are no predicted boxes with high quality at the earlier iteration in training. 

\begin{table}[t]
    \centering
    \small
    \caption{Ablation study on sample ratio for each target box type. The sample number is $128$. Our default setting is presented in the last row. GT refers to ground truth.}
    \begin{tabular}{c|c|c|ccc}
        \toprule
        Anchor & GT & Regressed Box & $AP$ & ${AP}_{50}$ & ${AP}_{75}$ \\ \midrule
        1.0  & 0.0  & 0.0  & 24.5 & 42.7 & 25.3  \\ \midrule
        0.0  & 1.0  & 0.0  & 32.4 & 51.8 & \textbf{33.9}  \\ \midrule
        0.0  & 0.0  & 1.0  & 31.9 & 51.4 & 33.4  \\ \midrule
        0.05 & 0.05 & 0.9  &\textbf{32.5} & \textbf{51.9} & 33.8  \\ \bottomrule
    \end{tabular}
    \label{Tab:sample_ratio}
\end{table}

\textbf{Inference Time} Table~\ref{Tab:FPS} summarizes an ablation study on inference time with different scales in Nvidia V100 GPU. Clearly, in the scale of 800 and 12 training epochs, PointINS gives faster inference speed among all compared method. In performance, it exhibits a notable gain compared to other PoI-based Polarmask (32.5 vs. 29.1) and MEInst (32.5 vs. 30.3), only inferior to inferior to RoI-based Mask R-CNN (32.5 vs. 34.4). However, when using data augmentation and longer training time (72 epochs), our method holds its inference speed advantage and gives competitive performance compared with Mask R-CNN (36.7 vs. 36.8), as shown in Table~\ref{Tab:overall}. Additionally, when the input scale decreases (\textit{e.g.}, 400), our model still achieves the practical performance (25.8) at real-time speed (27.3 fps). It indicates that PointINS can not only achieve high performance in mask AP, but also can be applied to real-time applications.
\begin{table}[t]
    \centering
    \small
    \caption{Ablation study on the inference time with different scales. All experiments are conducted with ResNet50 backbone.}
    \begin{tabular}{c|c|ccc|c}
        \toprule
        Scale & Method & $AP$ & ${AP}_{50}$ & ${AP}_{75}$ & FPS \\ \midrule
        
        800  & PolarMask~\cite{xie2020polarmask}  & 29.1  & 49.5 & 29.7 & 17.2  \\ \midrule
        800  & MEInst~\cite{zhang2020MEInst}  & 30.3  & 53.0 & 31.1 & 17.6  \\ \midrule
        800  & \multirow{3}{*}{PointINS}  & 32.5  & 51.9 & 33.8 & 18.2  \\ \cline{1-1} \cline{3-6}
        600  &   & 30.3  & 49.4 & 31.3 & 22.4  \\ \cline{1-1} \cline{3-6}
        400  &   & 25.8  & 44.7 & 26.4 & \textbf{27.3}  \\ \midrule
        800  & Mask R-CNN~\cite{DBLP:conf/iccv/HeGDG17} & \textbf{34.4} & \textbf{55.1} & \textbf{36.7} & 16.1 \\ 
        \bottomrule
    \end{tabular}
    \label{Tab:FPS}
\end{table}

\subsubsection{Comparison with State-of-the-Art Methods}
\textbf{Comparison among point-based methods}
We have implemented our method based on RetinaNet~\cite{lin2017focal} and FCOS~\cite{tian2019fcos}, and compare them with three state-of-the-art point-based instance segmentation frameworks: TensorMask~\cite{chen2019tensormask}, PolarMask~\cite{xie2020polarmask}, and MEInst~\cite{zhang2020MEInst}, as described in Table~\ref{Tab:overall}. It is clear that our PointINS framework performs the best among all frameworks. In the following, We present a comparison with these approaches.

\begin{table*}[t]
    \centering
    \small
    \caption{Comparison with current point-based instance segmentation frameworks on test-dev split of COCO dataset. `Aug' refers to data augmentation, including multi-scale and random crop. \checkmark (or $\circ$) signifies that the network is trained with (or without) data augmentation. `Epoch' indicates the training time. 12 epochs is a baseline widely used in most previous work \cite{DBLP:conf/iccv/HeGDG17,DBLP:journals/corr/abs-1803-01534,lin2017focal,tian2019fcos}.}
    \begin{tabular}{c|c|c|c|ccc|ccc}
        \toprule
        \multicolumn{10}{c}{\textbf{Two-stage}} \\ \midrule
        method & backbone & aug & epochs & $AP$ & ${AP}_{50}$ & ${AP}_{75}$ & ${AP}_{S}$ & ${AP}_{M}$ & ${AP}_{L}$ \\ \midrule
        
        MNC~\cite{dai2016instance} & R-101-C4 & $\circ$ & 12 & 24.6 & 44.3& 24.8 & 4.7 & 25.9 &43.6  \\ \midrule
        FCIS~\cite{li2016fully} & R-101-C5-dilated & $\circ$ & 12 & 29.2 & 49.5 & - & 7.1 & 31.3 & 50.0 \\ \midrule
        \multirow{3}*{Mask R-CNN~\cite{DBLP:conf/iccv/HeGDG17}} 
        & R-50-FPN & \checkmark & 72 & 36.8 & 59.2 & 39.3 & 17.1 & 38.7 & 52.1\\
        & R-101-FPN & \checkmark & 72 & 38.3 & 61.2 & 40.8 & 18.2 & 40.6 & 54.1\\
        & X-101-FPN & $\circ$ & 12 & 37.1 & 60.0 & 39.4 & 16.9 & 39.9 & 53.5 \\ \midrule

        \multicolumn{10}{c}{\textbf{One-stage (Segmentation-based)}} \\ \midrule
        method & backbone & aug & epochs & $AP$ & ${AP}_{50}$ & ${AP}_{75}$ & ${AP}_{S}$ & ${AP}_{M}$ & ${AP}_{L}$ \\ \midrule
        \multirow{2}*{YoLACT \cite{bolya-iccv2019}} & R-50-FPN & \checkmark & 45 & 28.2 & 46.6 & 29.2 & 9.2 & 29.3 & 44.8 \\ 
        & R-101-FPN & \checkmark & 45 & 31.2 & 50.6 & 32.8 & 12.1 & 33.3 & 47.1 \\  \midrule

        SOLO \cite{bolya-iccv2019} & R-101-FPN & \checkmark & 36 & 37.8 & 59.5 & 40.4 & 16.4 & 40.6 & 54.2 \\ \midrule
        EmbedMask \cite{ying2019embedmask} & R-101-FPN & \checkmark & 36 & 37.7 & 59.1 & 40.3 & 17.9 & 40.4 & 53.0 \\ \midrule

        \multicolumn{10}{c}{\textbf{One-stage (Point-based)}} \\ \midrule
        method & backbone & aug & epochs & $AP$ & ${AP}_{50}$ & ${AP}_{75}$ & ${AP}_{S}$ & ${AP}_{M}$ & ${AP}_{L}$ \\ \midrule
        ExtremeNet~\cite{zhou2019bottom} & Hourglass-101 & \checkmark & 100 & 18.9 & 44.5 & 13.7 & 10.4 & 20.4 & 28.3 \\ \midrule
        \multirow{2}*{TensorMask \cite{chen2019tensormask}} & R-50-FPN & \checkmark & 72  & 35.4 & 57.2 & 37.3 & 16.3 & 36.8 & 49.3  \\ 
        & R-101-FPN & \checkmark & 72 & 37.1 & 59.3 & 39.4 & 17.4 & 39.1 & 51.6 \\ \midrule
        
        \multirow{4}*{PolarMask \cite{xie2020polarmask}} & R-50-FPN & $\circ$ & 12 & 29.1 & 49.5 & 29.7 & 12.6 & 31.8 & 42.3\\ 
        & R-101-FPN & $\circ$ & 12 & 30.4 & 51.1 & 31.2 & 13.5 & 33.5 & 43.9 \\
        & R-101-FPN & $\checkmark$ & 24 & 32.1 & 53.7 & 33.1 & 14.7 & 33.8 &45.3 \\ 
        & X-101-FPN-DCN & $\checkmark$ & 24 & 36.2 & 59.4 & 37.7 & 17.8 & 37.7 & 51.5 \\ \midrule 

        \multirow{4}*{MEInst \cite{zhang2020MEInst}} & R-101-FPN & $\circ$ & 12 & 33.0 & 56.4 & 34.0 & 15.2 & 35.3 & 46.3 \\
        & R-101-FPN & $\checkmark$ & 36 & 33.9 & 56.2 & 35.4 & 19.8 & 36.1 &42.3 \\ 
        & X-101-FPN-DCN & $\checkmark$ & 36 & 38.2 & 61.7 & 40.4 & \textbf{22.6} & 40.0 & 49.3 \\ \midrule 
        
        \multirow{4}*{RetinaNet + ours} & R-50-FPN & $\circ$ & 12 & 32.2 & 51.6 & 33.4 & 13.4 & 34.4 & 48.4 \\
        & R-101-FPN & $\circ$ & 12 & 33.5 & 53.6 & 34.1 & 14.2 & 35.3 & 49.0\\ 
        & R-50-FPN & \checkmark & 72 & 36.2 & 58.1 & 37.8 & 16.5 & 37.7 & 50.5\\
        & R-101-FPN & \checkmark & 72  & 37.9 & 60.1 & 39.7 & 17.8& 40.1 & 52.1\\ \midrule
        \multirow{4}*{FCOS + ours} & R-50-FPN & $\circ$ & 12 & 33.4 & 53.7 & 35.2 & 14.8 & 36.3 & 48.8\\ 
        & R-101-FPN & $\circ$ & 12 & 34.5 & 54.5 & 36.6 & 15.4 & 37.0 & 49.4\\
        & R-50-FPN & \checkmark & 72 & 36.7 & 58.2 & 38.1 & 16.8 & 38.0 & 50.9\\ 
        & R-101-FPN & \checkmark & 72 & 38.3 & 60.3 & 40.0 & 18.1 & 40.3 & 52.4\\
        & X-101-FPN-DCN & \checkmark & 72 & \textbf{42.0} & \textbf{63.7} & \textbf{44.3} & 21.0 & \textbf{43.5} & \textbf{55.9}\\ \bottomrule
    \end{tabular}
    \label{Tab:overall}
\end{table*}

PolarMask~\cite{xie2020polarmask} uses polar representation as an approximation to the instance mask. 
With 12 epochs of training, PolarMask achieves 29.1 mAP with R-50-FPN backbone, and 30.4 mAP with R-101-FPN. With RetinaNet, our PointINS achieves 32.2 mAP, with an improvement of 3.1 points. The margin is even larger when we use FCOS as the detection base, we achieve 33.4 and 34.5 with R-50-FPN and R-101-FPN backbones, respectively. Note that the polar representation is an approximation to the ground truth, which intrinsically leads to the precision loss. In contrast, our method uses a pixel-wise mask and does not suffer from this problem. Our method with ResNet50 with only single-scale $1\times$ training still outperforms the method with ResNet101 under multi-scale $2\times$ training. $1\times$ and $2\times$ training mean training the network in 12 and 24 epochs.

MEInst~\cite{zhang2020MEInst} encodes a mask into a compact representation by principal component analysis (PCA). The network responds to generate feature encodings and finally decodes them to instance masks. Like the polar representation in PolarMask, this representation is coarser than the original ground truth, thus affecting the final segmentation quality. With the ResNet-101 backbone, MEInst only obtains the performance of approximately 33.0, which is lower than ours with only ResnetNet-50 backbone. In particular, large objects in MEInst lead to poorer performance because the compact representation misses too many details of large objects. 

Compared to the Tensormask~\cite{chen2019tensormask}, we achieve an improvement of 0.8 ($37.1 \rightarrow 37.9$) and 1.2 ($37.1 \rightarrow 38.3$) with RatinaNet and FCOS, respectively. 
The cause of our success is twofold. First, the instance-agnostic feature we use is more informative and robust than the aligned representation proposed by Tensormask. Second, we propose the instance-aware module, which allows our framework to capture accurate masks for the predicted instances. 

Note that by applying data augmentation and training for a longer time, our performance further improves significantly. By training for as long as 72 epochs and using data augmentation, our model with RetinaNet achieves 37.9 mAP with R-101-FPN backbone, while with FCOS it achieves 38.3 mAP. For a fair comparison with the previous winner, TensorMask~\cite{chen2019tensormask}, we use the same hyperparameter setting. 

\textbf{Comparison with State-of-the-Arts Methods}
Table~\ref{Tab:overall} also demonstrates that PointINS can obtain the comparable performance to Mask R-CNN~\cite{DBLP:conf/iccv/HeGDG17} and SOLO~\cite{wang2019solo}) by only using a single-point feature. This demonstrates that PointINS is a promising method that does not involve RoI feature extraction, and the point feature can be further improved with more elaborate designs.

\begin{table}[t]
    \centering
    \small
    \caption{The associated detection performance of the Tensormask and PointINS with ResNet50 backbone. All the experiemnts are conducted with data augmentation and 6$\times$ training scheme.}
    \begin{tabular}{c|c|c|ccc}
        \toprule
        method & backbone & type & $AP$ & ${AP}_{50}$ & ${AP}_{75}$ \\ \midrule
        \multirow{2}*{TensorMask} & \multirow{2}*{R-50-FPN} & box & 41.6 & 61.0 & 45.1  \\ 
        & & mask & 35.4 & 57.2 & 37.3 \\ \midrule
        \multirow{2}*{RetinaNet + ours} & \multirow{2}*{R-50-FPN} & box & 41.8 & 61.2 & 45.3 \\
        & & mask & \textbf{36.2} & 58.1 & 37.8 \\\midrule
		\multirow{2}*{FCOS + ours} & \multirow{2}*{R-50-FPN} & box & 42.3 & 61.7 & 45.9 \\
        & & mask & \textbf{36.7} & 58.2 & 38.1 \\\bottomrule
    \end{tabular}
    \label{Tab:det}
\end{table}

\subsubsection{Correlation to Object detection.}
Table~\ref{Tab:det} presents the associated detection performance of Tensormask and our PointINS framework on COCO test-dev set. Initially, the mask performance (36.2) of our RetinaNet-based PointINS is higher than that (35.4) of Tensormask with approximately the same box quality (41.6 vs 41.8). This shows that the mask improvement of the pointINS results from our designed modules, including instance-agnostic feature generation and instance-aware convolution. Both help our model capture more robust and aligned features for each specific box. Furthermore, the mask quality obtains consistent improvement with an effective one-stage detector such as FCOS.

\begin{figure*}[t]
\begin{center}
\includegraphics[height=8.3cm, width=1\linewidth]{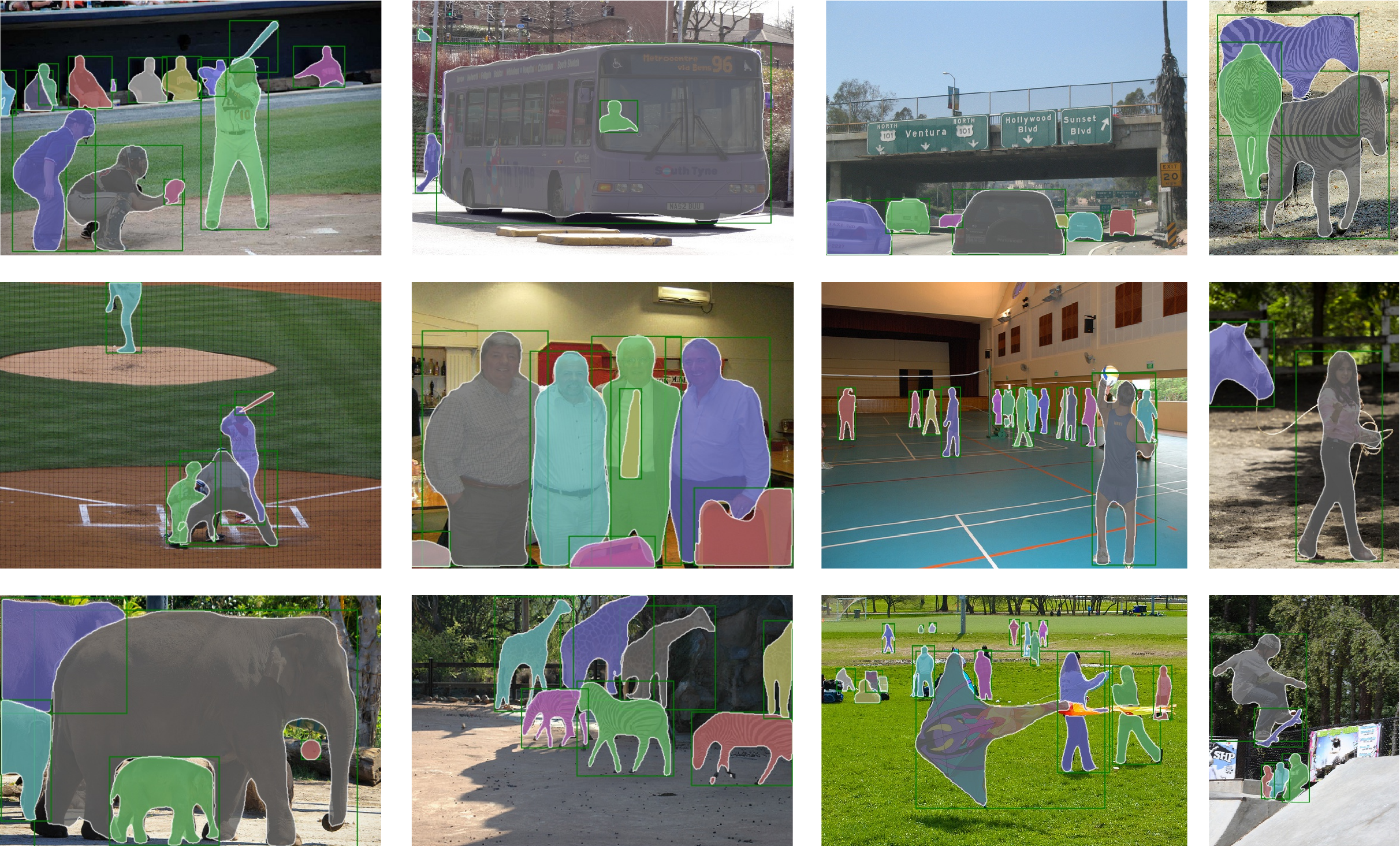}
\end{center}
\caption{Visualization of RetinaNet-based PointINS with ResNet101 on COCO dataset under $1\times$ (12 epochs) training. For all images, we set the mask confidence threshold to 0.5.}
\label{fig:visualization}
\end{figure*}

\subsubsection{Visualization}
Figure~\ref{fig:visualization} shows a visualization of our predicted instances. Notably, our approach can easily detect small objects, as our employed dense one-stage detector excels at this. This is also verified in Table~\ref{Tab:overall} with better ${AP}_{s}$ than SOLO~\cite{wang2019solo}. For large objects, the boundary is slightly worse than that of small ones. This signifies that a single-point feature in our PointINS compressed detailed information for large regions due to the pyramid features used in object detection and instance segmentation frameworks. Like RetinaNet and FCOS, large objects are detected with P6 or P7 features, which are highly abstract to estimate the location of large objects. However, they are not suitable for pixel-level instance segmentation because they lose details by downsampling. Longer training time and data augmentation can relieve this problem, as illustrated in Table~\ref{Tab:overall}.

\subsection{Experiments on Cityscapes}
\textbf{Dataset and Metrics}
The Cityscapes~\cite{cordts2016cityscapes} dataset contains street scenes captured by car-mounted cameras. There are 2,975 training, 500 validation, and 1,525 testing images with fine annotations. Here, we report our results on both validation and test subset. Eight semantic classes are annotated with instance masks, and each image had a size of $1024 \times 2048$. We evaluate results based on AP and AP50.

\textbf{Hyper-parameters}
We use images with shorter edges randomly sampled from $\{800, 1024\}$ for training, and images with a shorter edge length of $1024$ for inference. We train our model with a learning rate of 0.0005 for 18k iterations and with a rate of 0.00005 for other 6k iterations. Eight images (1 image per GPU) are in one image batch.

\begin{table*}[t]
\centering
\small
\caption{Resuts on cityscapes dataset. All experiments are conducted with ResNet50. "Mcycle" is short for motorcycle.}
\begin{tabular}{c|c|cc|cccccccc}
\toprule
Method & ${AP}_{val}$  & $AP$ & ${AP}_{50}$ & person & rider & car & truck & bus & train & mcycle & bicycle \\ \midrule
SGN~\cite{liu2017sgn} & 29.2 & 25.0 & 44.9 & 21.8 & 20.1 & 39.4 & \textbf{24.8} & \textbf{33.2} & \textbf{30.8} & 17.1 & 12.4 \\ \midrule
Mask R-CNN~\cite{DBLP:conf/iccv/HeGDG17} & 31.5 & 26.2 &  49.9 & 30.5 & 23.7 & 46.9 & 22.8 & 32.2 & 18.6 & 19.1 & 16.0 \\ \midrule
Ours & \textbf{32.4} & \textbf{27.3} & \textbf{50.5} & \textbf{34.1} & \textbf{26.1} & \textbf{51.9} & 20.9 & 30.9 & 15.9 & \textbf{20.5} & \textbf{19.2} \\ \bottomrule
\end{tabular}
\label{Tab:cityscapes}
\end{table*}
\textbf{Results}
Table~\ref{Tab:cityscapes} presents the results of our PointINS framework with the FCOS-Res50 backbone. PointINS outperforms RoI and segmentation-based methods with a simple yet efficient structure.

\section{Conclusion}
In this study, we introduce PointINS to convert current dense one-stage detectors for instance segmentation by a single-point feature. 
The core module is instance-aware convolution, ensuring that single-point features are sufficiently expressive for instance masks.
This module decomposes a single point feature into two tractable modules: an instance-agnostic feature and instance-aware weight generation modules. This design ensures that high-dimensional instance-aware mask features of several positive points are aligned (by convolving the template with dynamic weights) and are then used for mask prediction.
The experiments show that the proposed framework achieves competitive accuracy and speed among point-based frameworks. Furthermore, the proposed framework is comparable to the popular Mask R-CNN framework. 

In the future, we will explore new dynamic weight generation methods to make a single-point feature robust. In addition, we will study how to apply this concept to other instance-based downstream tasks like panoptic segmentation~\cite{kirillov2019panoptic,xiong2019upsnet,li2019attention} or visual relationships prediction~\cite{kuznetsova2018open} or dense captioning~\cite{johnson2016densecap}.

{
  \small
  \bibliographystyle{ieee}
  \bibliography{egbib}
}
\end{document}